\documentclass[runningheads]{llncs}

\usepackage[utf8]{inputenc}
\DeclareUnicodeCharacter{2212}{-}
\usepackage{graphicx}
\usepackage{amsmath,amssymb} % define this before the line numbering.
\usepackage{comment}
\usepackage{subcaption}
\usepackage{hyperref}
\usepackage{etoolbox,siunitx}
\usepackage{pifont}
\usepackage{pgf}
\usepackage{pgfplots}
\pgfplotsset{compat=1.15}

\robustify\bfseries
\usepackage{soul}

\usepackage{booktabs}
\usepackage{color}

\newcommand{\etal}{\emph{et~al}.}

\newcommand{\pcent}[1]{\SI{#1}{\percent}}

\newcommand*{\convmap}{\mbox{\emph{ConvMap}} }
\newcommand*{\convmaps}{\mbox{\emph{ConvMaps}} }

\newif\ifreview
\reviewfalse

\ifreview
   \usepackage{lineno}

   \linenumbers
\fi

\makeatletter
\newcommand\footnoteref[1]{\protected@xdef\@thefnmark{\ref{#1}}\@footnotemark}
\makeatother

\begin{document}
   \pagestyle{headings}
   \mainmatter
   %!TEX root = main.tex

% Insert your submission number here
\def\SubNumber{86}

% Choose one track by uncommenting one of the following lines
\def\GCPRTrack{Regular Track}
% \def\GCPRTrack{Track: Computer vision systems and applications}
% \def\GCPRTrack{Track: Pattern recognition in the life and natural sciences}
% \def\GCPRTrack{Track: Photogrammetry and remote sensing}
% \def\GCPRTrack{Track: Robot vision}
% \def\GCPRTrack{Track: DAGM Young Researcher Forum}

% Replace with your title
% \title{End-to-end Learning of a Fisher Vector Encoding for Part Features in Fine-grained Recognition}
\title{End-to-end Learning of Fisher Vector Encodings for Part Features in Fine-grained Recognition}
\titlerunning{End-to-end FVE for Part Features in Fine-grained Recognition}
% You can use \thanks for acknowledgment. Do not add any acknowledgment to the draft
% version that is used for the review process.
%\title{Title\thanks{XXX}}

\ifreview
   % ANONYMOUS SUBMISSION FOR REVIEW
   % DO NOT MODIFY these for the draft version that is used for the review process.
   \titlerunning{DAGM GCPR 2021 Submission \SubNumber{}. CONFIDENTIAL REVIEW COPY.}
   \authorrunning{DAGM GCPR 2021 Submission \SubNumber{}. CONFIDENTIAL REVIEW COPY.}
   \author{DAGM GCPR 2021 - \GCPRTrack{}}
   \institute{Paper ID \SubNumber}
\else
   % CAMERA READY SUBMISSION
   %\titlerunning{Abbreviated paper title}
   % If the paper title is too long for the running head, you can set
   % an abbreviated paper title here

   \author{
      Dimitri Korsch\inst{1} \and
      Paul Bodesheim\inst{1} \and
      Joachim Denzler\inst{1,2,3}
   }

   \authorrunning{D. Korsch et al.}
   % First names are abbreviated in the running head.
   % If there are more than two authors, 'et al.' is used.

   \institute{
      Computer Vision Group, Friedrich-Schiller-University Jena, Germany \and
      Michael Stifel Center Jena for Data-Driven and Simulation Science, Jena, Germany \and
      German Aerospace Center (DLR), Institute for Data Science, Jena, Germany
   }
\fi

\maketitle              % typeset the header of the contribution

   %!TEX root = ../main.tex

\begin{abstract}
Part-based approaches for fine-grained recognition do not show the expected performance gain over global methods, although explicitly focusing on small details that are relevant for distinguishing highly similar classes.
We assume that part-based methods suffer from a missing representation of local features, which is invariant to the order of parts and can handle a varying number of visible parts appropriately.
The order of parts is artificial and often only given by ground-truth annotations, whereas viewpoint variations and occlusions result in not observable parts.
Therefore, we propose integrating a Fisher vector encoding of part features into convolutional neural networks.
The parameters for this encoding are estimated by an online EM algorithm jointly with those of the neural network and are more precise than the estimates of previous works.
Our approach improves state-of-the-art accuracies for three bird species classification datasets.% on CUB-200-2011 from 90.3\% to 91.1\%, on NA-Birds from 89.2\% to 90.3\%, and on Birdsnap from 84.3\% to 84.9\%.
% \todo{145 of max. 150 words!}
\end{abstract}

   %!TEX root = ../main.tex

\section{Introduction}
\label{sec:intro}

\begin{figure}[t]
	\centering
	\includegraphics[width=0.85\textwidth]{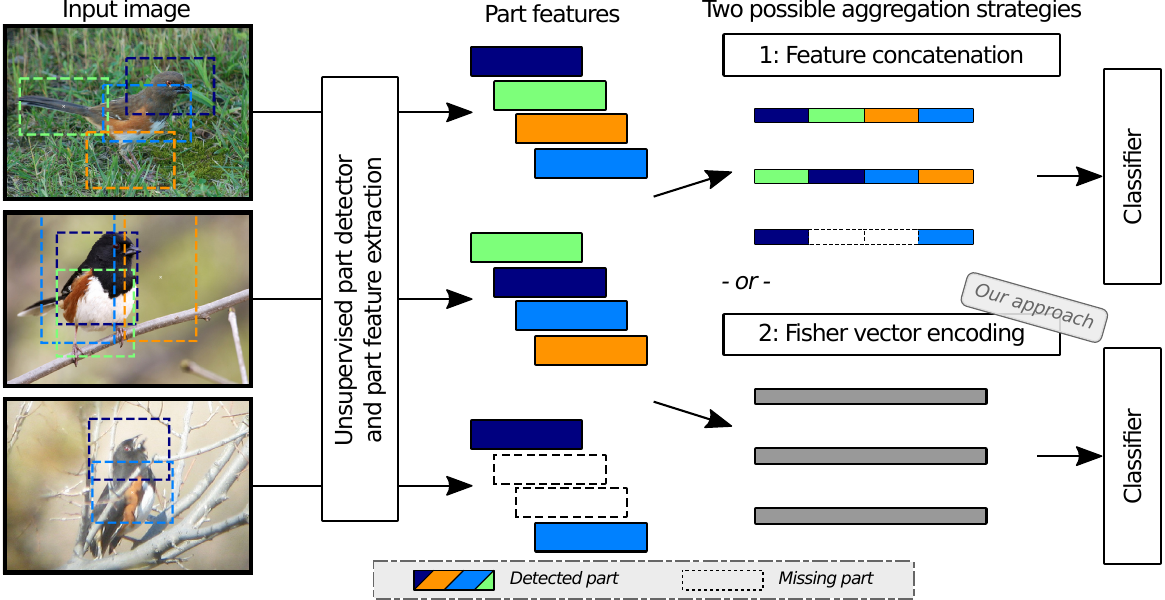}
	\caption{When some parts can not be detected because they are not visible, the resulting gaps of missing features need to be filled when features are concatenated. Furthermore, the semantic meaning of extracted part features is not clear when applying unsupervised part detection algorithms that do not preserve a consistent order of part features. With our approach, we compute a Fisher vector encoding as a unified representation of fixed length for an arbitrary number of unordered part features, which can be used by any type of classifier, including simple linear classifiers and fully-connected layers in a deep neural network.}
	\label{fig:teaser}
	\vspace{-0.5cm}
\end{figure}

Part- or attention-based approaches~\cite{Fu_2017_CVPR,ge2019weakly,he2019and,zhang2019unsupervised,zhang2019learning,zheng2017learning} are common choices for fine-grained visual categorization because they explicitly focus on small details that are relevant for distinguishing highly similar classes, e.g., different bird species.
Quite surprisingly, methods that perform the categorization with global image features~\cite{Cui_2018_CVPR_large,lin2015bilinear,Simon17_GOP,Simon19:Implicit,touvron2019fixing,zheng2019learning} also achieve excellent results.
It is hard to tell from the empirical results reported in the literature which general approach (global or part-based) is superior, given that all of them show comparable results in terms of recognition performance.
We hypothesize that part-based algorithms cannot exploit their full potential due to the problems that arise from the initial detection of parts, especially regarding a unified representation of individual part features after the detection.
Since learning individual part detectors requires part annotations and thus additional, time-consuming efforts by domain experts, methods for unsupervised part detection have been developed that already obtain remarkable classification results~\cite{ge2019weakly,Korsch19_CSPARTS,zhang2019learning}.
However, unsupervised part detection faces various challenges, such as missing parts caused by different types of occlusions and parts with ambiguous semantic meaning, as shown in Fig.~\ref{fig:teaser}.
Hence, it remains unclear whether detected parts are reasonable and semantically consistent.
Furthermore, part-based classifiers usually require a fixed number of parts to be determined for each image and a pre-defined order of the extracted part features. %, which are then handed over to the classifier.
These are strong restrictions for the application, especially when considering varying poses and viewpoints that lead to hidden parts.
We believe that a common way for representing a \emph{varying number of unordered part features} obtained from every single image is rarely used in the context of fine-grained classification.

Fisher vector encoding (FVE)~\cite{perronnin2007fisher,Perronnin10:ITF} is a well-known feature transformation method typically applied to local descriptors of key points.
When applied to part features, it allows for computing a unified representation of fixed length for each image, independent of the number of detected parts (Fig.~\ref{fig:teaser}).
Note that the problem of missing parts can also be observed for ground-truth part annotations of fine-grained bird species datasets~\cite{Birdsnap,NABirds,CUB_200_2011}.
Due to various poses and occlusions, some bird parts are not visible and cannot be annotated.
These datasets are often used to evaluate the performance of classifiers without taking the part detection into account, and this requires a proper gap-filling strategy for missing parts.
In contrast, no gap-filling is needed when using the FVE.
Furthermore, FVE allows for neglecting an artificial order of parts since there is usually no natural order of parts, and each part should be treated equally.

Applying FVE together with CNNs has been done differently in the past: either GMM and FVE are computed after training the CNN model~\cite{cimpoi2015deep,song2017locally,zhang2016picking} such that only FVE parameters are adapted to the CNN features and not vice versa, or the FVE has been realized as a trainable layer~\cite{arandjelovic2016netvlad,gong2014vlad_pooling,tang2018deep,wieschollek2017backpropagation}.
Although the latter allows for end-to-end training of GMM parameters, artificial constraints are required to obtain reasonable values (e.g., positive variances).
Furthermore, only the classification loss influences the mixture estimations, and there is no objective involved for modeling the data distribution correctly.
In contrast, we propose to realize the FVE as a differentiable feature transformation based on a GMM estimated with an iterative EM algorithm using mini-batch updates of the parameters.
First, the differentiable transformation allows end-to-end training of the feature extraction and classification weights.
Next, the FVE directly influences the feature extraction within the CNN such that the features adapt to the encoding.
Finally, since the parameters of the GMM are estimated with an EM algorithm instead of gradient descent, the resulting GMM describes the input data more precisely.

In our experiments, we show that the FVE outperforms other feature aggregation methods and that our approach estimates the GMM parameters more precisely than gradient descent methods.
Furthermore, our approach described in Sect.~\ref{sec:methods} improves state-of-the-art accuracies for three fine-grained bird species categorization datasets: CUB-200-2011\cite{CUB_200_2011} (from 90.3\% to 91.1\%), NA-Birds\cite{NABirds} (from 89.2\% to 90.3\%), and Birdsnap\cite{Birdsnap} (from 84.3\% to 84.9\%).

   %!TEX root = ../main.tex

\section{Related Work}

% We divide this section into three parts.
First, we discuss related work on FVE in the context of deep neural networks.
Some of these approaches either do not allow for learning the GMM parameters end-to-end but rather estimate parameters separately after neural network training.
Others treat GMM parameters as conventional network parameters learned without any clustering objective by artificially enforcing reasonable values for the mixture parameters.
Second, we review existing algorithms for iterative EM algorithms since we borrow ideas from these approaches. % for enabling end-to-end learning of GMM parameters together with CNN weights as part of our proposed FVE.
Surprisingly, none of these iterative approaches has been integrated in a CNN yet.
Third, we list current state-of-the-art techniques for fine-grained categorization, mainly focusing on part-based methods.
Some approaches~\cite{cimpoi2015deep,zhang2016picking} use FVE in fine-grained approaches, but all of them deploy the encoding in an offline manner, i.e., the FVE is used \emph{after} learning CNN parameters for feature extraction.

\subsection{Variants of Deep Fisher Vector Encoding (Deep FVE)}
\label{sub:related:deep_fisher_vector_encoding}

Simonyan~\etal~\cite{simonyan2013fve_layer} presented a Fisher vector layer for building deep networks.
They encode an input image or pre-extracted SIFT features in multiple layers with an FVE, but the entire network is trained greedily layer-by-layer and not end-to-end due to some restrictions.
Sydorov~\etal~\cite{sydorov2014deep} suggested another deep architecture that learns an FVE by updating GMM parameters based on gradients that are backpropagated from a Hinge loss of an SVM classifier.
Cimpoi~\etal~\cite{cimpoi2015deep} proposed a CNN together with an FVE for local CNN features, and the resulting feature representation is used by a one-vs-rest SVM.
Song~\etal~\cite{song2017locally} improved this approach, but still without end-to-end learning.

In contrast to the methods above, Wieshollek \etal~\cite{wieschollek2017backpropagation} and Tang \etal~\cite{tang2018deep} deploy the FVE directly in a neural network.
As a result, features are learned jointly with the classification and mixture model parameters.
However, although GMM parameters are estimated jointly with other network parameters using gradient descent, the training procedure has some drawbacks.
First, artificial constraints have to be applied to obtain reasonable mixture parameters, e.g., positive variances.
Next, due to the formulas for the FVE, the resulting gradients cause numerically unstable computations.
Finally, these approaches require a proper initialization of the mixture parameters, which implies the computation of the features of the entire dataset.
Performing such an initialization on large dataset results in an unreasonable computation and storage overhead.
% Same issues arise in the prominent NetVLAD~ approach.
Arandjelovic \etal~\cite{arandjelovic2016netvlad} employ a simplified version of the FVE, called Vector of Locally Aggregated Descriptors (VLAD)~\cite{jegou2010aggregating}, and rewrite its computation in terms of trainable network parameters that are optimized end-to-end via gradient descent.
However, for this NetVLAD-Layer~\cite{arandjelovic2016netvlad} as well as for~\cite{wieschollek2017backpropagation,tang2018deep}, it is arguable whether estimated GMM parameters are reasonable due to gradient-based optimization with back-propagation.
In contrast, our proposed method does not require any preliminary initialization since it uses an iterative EM algorithm to estimate the mixture parameters end-to-end, leading to meaningful cluster representations.

\subsection{EM Algorithms and GMM estimation}
\label{sub:related:em_algorithms_and_gmm_estimation}

The standard technique for estimating GMM parameters is the EM algorithm.
It is an iterative process of alternating between maximum likelihood estimation of mixture parameters and computing soft assignments of samples to the mixture components.
In the default setting, all samples are used in both steps, but this leads to an increased runtime for large-scale datasets.
To reduce the computational costs, one can only use a subset of samples for the parameter estimation or rely on existing online versions of the EM algorithm~\cite{OnlineEM:2009,neal1998view}.
For example, Cappé and Moulines~\cite{OnlineEM:2009} approximate the expectation over the entire dataset with an exponential moving average over batches of the data.
Based on this work, Chen \etal~\cite{chen2018stochastic} propose a variance reduction of the estimates in each step, which results in faster and more stable convergence.
A comparison of these algorithms is carried out by Karimi~\etal~\cite{karimi2019global}, who also introduce a new estimation algorithm. %and establish non-asymptotic convergence bounds for global convergence.
Further approaches~\cite{SequentialEM:2009,adaptiveGMM:2009,liang2009onlineEM} propose similar solutions with different applications and motivations for the iterative parameter update.

In our work, we employ the ideas of iterative parameter update coupled with a bias correction.
Furthermore, we demonstrate how to integrate these ideas in a neural network and estimate the parameters jointly with the network weights.

\subsection{Fine-grained visual categorization}
\label{sub:related:fine_grained_visual_categorization}

In the literature, two main directions can be observed for fine-grained recognition: global and part-based methods.
Global methods use the input image as a whole and employ clever strategies for pre-training~\cite{Cui_2018_CVPR_large}, augmentation~\cite{touvron2019fixing,krause2016unreasonable}, or pooling~\cite{lin2015bilinear,Simon17_GOP,Simon19:Implicit,zheng2019learning}.
In contrast, part- or attention-based approaches apply sophisticated detection techniques to determine interesting image regions and to extract detailed local features from these patches.
It results in part features as an additional source of information for boosting the classification performance.

He \etal~\cite{he2019and} propose a reinforcement learning method for estimating how many and which image regions are helpful for the classification.
They use multi-scale image representations for localizing the object and afterward estimate discriminative part regions.
Ge \etal~\cite{ge2019weakly} present the current state-of-the-art approach on the CUB-200-2011 dataset.
Based on weakly supervised instance detection and segmentations, part proposals are generated and constrained by a part model.
The final classification is performed with a stacked LSTM classifier and context encoding.
The method of Zhang \etal~\cite{zhang2019learning} also yields good results on the CUB-200-2011 and the NA-Birds dataset.
Expert models arranged in multiple stages predict class assignments and attention maps that the final expert uses to crop the image and refine the observed data.
Finally, a gating network is used to weigh the decisions of the individual experts.

Compared to the previous approaches, we use a different part detection method described at the beginning of the next section before presenting the details of our proposed FVE for part features.

   %!TEX root = ../main.tex

\section{Fisher Vector Encoding (FVE) of Part Features}
\label{sec:methods}

In this section, we present our approach for an FVE of part features, which allows for joint end-to-end learning of all parameters, i.e., the parameters of the underlying GMM and the parameters of the CNN that computes the part features.
It can be applied to any set of extracted parts from an image.
Hence it is possible to combine it with different part detection algorithms.
In this paper, we use the code\footnote{\url{https://github.com/cvjena/l1_parts}} for a part detection method provided by Korsch \etal~\cite{Korsch19_CSPARTS}.
The authors use an initial classification of the entire input image to identify features used for this classification.
Then, the pixels in the receptive field of these features are clustered and divided into candidate regions.
Bounding boxes are estimated around these regions and used as parts in the final part-based classification.

\begin{figure}[t]
	\centering
	\includegraphics[width=.85\textwidth]{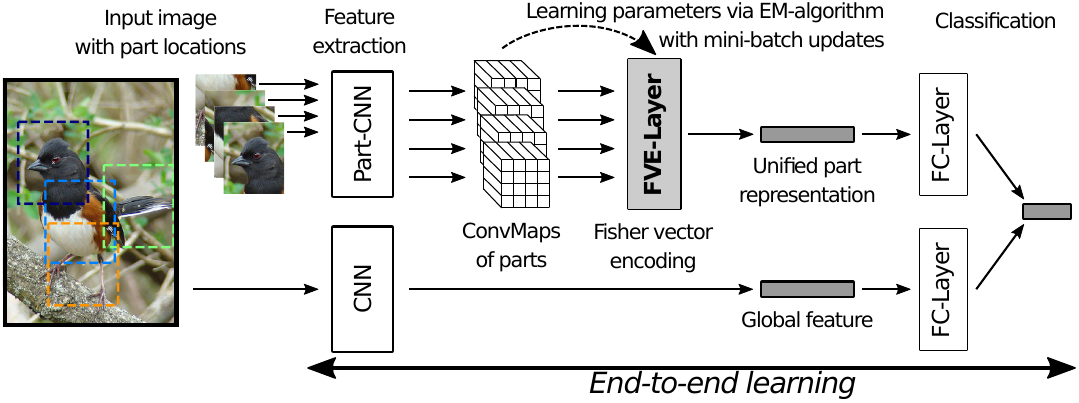}
	\caption{Overview of our proposed method. During training, we estimate the parameters of the GMM that leads to the FVE using an EM-algorithm with mini-batch updates described in Sect.~\ref{sub:estimation_of_parameters}. The resulting FVE-Layer, which is explained in Sect.~\ref{sub:training_with_the_fve_layer}, can be integrated in any deep network architecture. We use this new layer for computing a unified part representation that aggregates local features extracted from \convmaps of a CNN. Our approach enables joint end-to-end learning of both CNN parameters and GMM parameters for the FVE.}
	\label{fig:method}
	\vspace{-0.5cm}
\end{figure}

As shown in Figure~\ref{fig:method}, given a set of parts specified by their corresponding image regions, we propose the computation of a set of local features for each part with a CNN.
We denote the output of a CNN as a \convmap $C \in \mathbb{R}^{H \times W \times D}$, that consists of $N=H \cdot W$ local $D$-dimensional features $\mathcal{X} = \{\vec{x}_1, \dots, \vec{x}_N\}$.
Usually, CNNs contain \emph{global average pooling (GAP)} to reduce $\mathcal{X}$ to a single feature representation: $GAP(\mathcal{X}) = \frac{1}{N}\sum_{n=1}^N \vec{x}_n$.
Common part-based approaches~\cite{ge2019weakly,Korsch19_CSPARTS,yang2018_NTSNet,yang2021rerank} extract a set of \convmaps $\mathcal{C}=\{C_1,\dots,C_T\}$ from a single image by processing $T$ image regions and use GAP for each \convmap followed by concatenation of the resulting $T$ part features.
In contrast, we use FVE to transform the set of \convmaps into a single feature: $FVE(\mathcal{C}) = \vec{f} \in \mathbb{R}^{\hat{D}}$.

%  followed by an FVE of these part features to obtain a unified part representation of fixed length independent of the number of detected parts.
% Fig.~\ref{fig:method} visualizes our approach.
% Note that any CNN architecture can be used as a backbone network for computing part features, even with pre-trained weights that are fine-tuned when estimating the GMM parameters for the FVE.
% In order to characterize our proposed FVE that is learned end-to-end together with the network parameters, we present the mathematical formulation of the
% underlying mixture model.

\subsection{Fisher Vector Encoding}
\label{sub:fisher_vector_encoding}
First, we assume that the CNN computes local features \emph{i.i.d.} from an input image.
We further assume that all local descriptors of the extracted \convmaps $\mathcal{C}$ come from the same distribution with density function $p(\vec{x}_{n,t})$, represented by a finite mixture model with $K$ components: $p(\vec{x}_{n,t}|\Theta) = \sum_{k=1}^{K}\alpha_k p_k(\vec{x}_{n,t}|\theta_k)$ with mixture weights $\alpha_k$ that add up to 1 ($\sum_{k=1}^{K}\alpha_k = 1$), and model parameters $\Theta~=~\{\alpha_1,\theta_1,\dots,\alpha_K,\theta_K\}$.
Without any prior knowledge, let the density function of each component be a Gaussian distribution with mean vector $\vec\mu_k$ and diagonal covariance matrix $\vec\sigma_k$: $p_k(\vec{x}_{n,t}|\theta_k) = \mathcal{N}(\vec{x}_{n,t}|\vec\mu_k,\vec\sigma_k)$ leading to a Gaussian Mixture Model (GMM) with parameters $\Theta = \{\alpha_1,\vec\mu_1,\vec\sigma_1\dots\alpha_K,\vec\mu_K,\vec\sigma_K\}$.

Following Jaakola and Haussler~\cite{jaakkola1999exploiting}, and Perronnin \etal~\cite{perronnin2007fisher,Perronnin10:ITF}, the FVE is derived by considering the gradients of the log-likelihood with respect to the GMM parameters $\Theta$ and assuming independence of the part features: $\mathcal{F}_\Theta(\mathcal{C})=\sum_{n,t=1}^{N,T}\nabla_{\Theta}\log p(\vec{x}_{n,t}|\Theta)$.
These gradients, also called Fisher scores, describe how parameters contribute to the process of generating a particular feature.
We use the approximated normalized Fisher scores introduced by \cite{perronnin2007fisher,Perronnin10:ITF}:

\begin{align}
	\mathcal{F}_{\vec{\mu}_k}(\mathcal{C})
		&= \frac{1}{\sqrt{N \alpha_k}}\sum_{n,t=1}^{N,T} w_{n,t,k} \left(\dfrac{\vec{x}_{n,t} - \vec{\mu}_k}{\vec{\sigma}_k}\right)\quad,\label{eq:fisher_score_mean}
		\\
	\mathcal{F}_{\vec{\sigma}_k}(\mathcal{C})
		&= \frac{1}{\sqrt{2 N \alpha_k}}\sum_{n,t=1}^{N,T} w_{n,t,k} \left(\dfrac{(\vec{x}_{n,t} - \vec{\mu}_k)^2}{\vec{\sigma}_k^2} - 1\right) \label{eq:fisher_score_variance}
\end{align}
as feature encoding with $w_{n,t,k}=\frac{\alpha_k p(\vec{x}_{n,t}|\theta_k)}{\sum_{l=1}^{K}\alpha_l p(\vec{x}_{n,t}|\theta_l)}$ denoting the soft assignment of the feature $\vec{x}_{n,t}$ to a component $k$.

Finally, these scores can be computed for all parameters $\vec\mu_k$ and $\vec\sigma_k$ of the estimated GMM.
We use the concatenation of the scores as FVE of the part features, which results in a unified representation of dimension $2KD$ that is independent of the number of part features $T$.

% subsection fisher_vector_encoding (end)

\subsection{Estimation of the Mixture Model Parameters}
\label{sub:estimation_of_parameters}
The computations of the FVE require a GMM, and we illustrate two ways for estimating its parameters from data jointly with the CNN parameters.
In our experiments, we use both methods and compare them by the classification accuracy and quality of the estimated GMM parameters.

\noindent\textbf{Gradient-based estimation}\quad
This idea is covered in different works~\cite{arandjelovic2016netvlad,tang2018deep,wieschollek2017backpropagation}: since all of the operations in Eqs.~\eqref{eq:fisher_score_mean} and~\eqref{eq:fisher_score_variance} are differentiable, it is straightforward to implement the FVE as a differentiable FVE-Layer such that the parameters~$\Theta$ are estimated via gradient descent.
However, the GMM constraints (positive variances and prior weights adding up to one) need to be enforced.
Wieschollek~\etal~\cite{wieschollek2017backpropagation} propose to model the variances $\sigma^2_k=\epsilon + \exp(s_k)$ and the mixture weights $\alpha_k=\frac{\mathrm{sigm}(a_k)}{\sum_j\mathrm{sigm}(a_j)}$ by estimating $s_k$ and $a_k$ instead of $\sigma^2_k$ and $\alpha_k$.

\noindent\textbf{Online-EM estimation}\quad
Different variants for an online EM algorithm can be found in the literature~\cite{SequentialEM:2009,OnlineEM:2009,chen2018stochastic,karimi2019global,liang2009onlineEM,neal1998view}.
The main idea is to approximate the expectations over the entire dataset with exponential moving averages~(EMAs) over batches of the data: $\Theta[t] = \lambda\cdot\Theta[t-1] + (1-\lambda) \cdot \Theta^{\mathrm{new}}$ with $\lambda \in (0, 1)$ and $[t]$ indicating the training step $t$.
We follow this approach and propose an FVE-Layer with online parameter estimation via EMAs.
It is worth mentioning that the parameters of the widely used batch-normalization layer~\cite{ioffe2015batch} are also estimated with EMAs.
However, our FVE-Layer differs from a batch-normalization layer in two ways.
First, we estimate a mixture of Gaussians and not only the mean and variance of the inputs.
Second, we encode the input according to Eqs.~\eqref{eq:fisher_score_mean} and \eqref{eq:fisher_score_variance} instead of whitening the inputs.
Additionally, we perform bias correction via $\hat{\Theta}[t] = \frac{\Theta[t]}{1-\lambda^t}$ since plain EMAs are biased estimators.
Similar bias correction is also done in the Adam optimizer~\cite{kingma2014adam}.

Finally, we have observed that some of these $D$-dimensional local feature vectors ($H{\cdot}W$ vectors for each of the $T$ parts) have low $L^2$-norm, especially if the corresponding receptive field mainly contains background pixels.
However, since we are only interested in using local features that exceed a certain activation level, i.e., that carry important information, we include an additional filtering step for the local feature vectors before both the estimation of the GMM parameters and the computation of the FVE.
We only use local features with an $L^2$-norm greater than the mean $L^2$-norm of all local features obtained from the same image.
We found that this filtering leads to more stable and balanced estimates for the GMM parameters during our experiments.

\subsection{Training with the FVE-Layer} % (fold)
\label{sub:training_with_the_fve_layer}

We implement the proposed FVE-Layer utilizing the calculations from Eqs.~\eqref{eq:fisher_score_mean} and~\eqref{eq:fisher_score_variance} as well as the online parameter estimation introduced in the previous section.
For end-to-end training of the CNN layers preceding the FVE-Layer, we estimate the gradients of the encoding w.r.t. the inputs similar to~\cite{wieschollek2017backpropagation}:

\begin{align}
\frac{\partial\mathcal{F}_{\mu_{k,d}}(\mathcal{X}_I)}{\partial{x_{n,d_*}}}
	&= \dfrac{1}{\sqrt{N_I\alpha_k}}\left[\frac{\partial{w_{n,k}}}{\partial{x_{n,d_*}}}\left(\dfrac{x_{n,d} - \mu_{k,d}}{\sigma_{k,d}}\right) + \delta_{d,d_*}\frac{w_{n,k}}{\sigma_{k,d_*}}\right] \quad,\\
\frac{\partial\mathcal{F}_{\sigma_{k,d}}(\mathcal{X}_I)}{\partial{x_{n,d_*}}}
	&= \dfrac{1}{\sqrt{2N_I\alpha_k}}\left[
	     \frac{\partial{w_{n,k}}}{\partial{x_{n,d_*}}}\left(\dfrac{(x_{n,d} - \mu_{k,d})^2}{(\sigma_{k,d})^2} - 1\right)\right. \nonumber \\
	     & \phantom{==\dfrac{1}{\sqrt{2N_I\alpha_k}}}\left.+\delta_{d,d_*}\frac{2 w_{n,k}(x_{n,d_*}-\mu_{k,d_*})}{(\sigma_{k,d_*})^2}\right]\quad.
\end{align}
In both equations, we use $\delta_{d,d_*}$ to denote the Kronecker delta being $1$ if $d=d_*$ and $0$ else, as well as the derivative of $w_{n,k}$ w.r.t. $x_{n,d_*}$ that is given by $
\frac{\partial{w_{n,k}}}{\partial{x_{n,d_*}}} = w_{n,k}\left(-\frac{(x_{n,d_*}-\mu_{k,d_*})}{(\sigma_{k,d_*})^2} + \sum_{\ell=1}^K w_{n,\ell}\frac{(x_{n,d_*}-\mu_{\ell,d_*})}{(\sigma_{\ell,d_*})^2} \right)$.
Further details for the derivation of these gradients can be found in the supplementary material.

Though these gradients are computed within the deep learning framework by the autograd functionality, it is important to mention that we observed some numerical instabilities during the training, especially with high dimensional mixture components.
% \todo{mention that these instabilties are also mainly observed for the gradient-based FVE? Dima: kann ich leider nicht mir sicherheit sagen, ob es eher beim Grad-based ist.}
To circumvent this issue, we perform an auxiliary classification on the inputs of the FVE-Layer, similar to Szegedy~\etal~\cite{szegedy2015going}.
The auxiliary classification branch consists of a global average pooling and a linear layer.
Finally, we combine the resulting auxiliary loss with the loss computed from the prediction on the encoded part features: $\mathcal{L}_{parts}=\beta \cdot \mathcal{L}_{aux} + (1-\beta) \cdot \mathcal{L}_{FVE}$.
We set $\beta$ to \num{0.5} and multiply it by \num{0.5} after \num{20} epochs, such that the effect of the auxiliary classification decreases over time.
Furthermore, we omit the auxiliary branch for the final classification and perform the part classification entirely on the features encoded by the FVE.

The final loss, consisting of the losses computed from the global and the part predictions, is computed in a similar way: $\mathcal{L}_{final}=\frac{1}{2}\left(\mathcal{L}_{parts} + \mathcal{L}_{global}\right)$. For cross-entropy, this combination is equivalent to computing the final prediction as a geometric mean of the class probabilities or the arithmetic mean of the normalized log-likelihoods (see supplementary material for more details).
% \todo{mean of the likelihoods? which likelihoods? you mean individual (part and global) predictions? Dima: ja, also die Klassenwahrscheinlichkeiten die nach dem Softmax kommen}

% \begin{enumerate}
% 	\item FVE-Layer and its gradients w.r.t. inputs
% 	\item auxiliary loss for regularization (due to complex Gradients of the FVE-Layer)
% 	\item part and global loss combination
% \end{enumerate}

% subsection training_with_the_fve_layer (end)

   %!TEX root = ../main.tex

\section{Experimental results}
\label{sec:results}

\subsection{Datasets}
\label{sub:res:datasets}
We evaluate our method on widely used datasets for fine-grained categorization.
First, we use three datasets for bird species recognition: \emph{CUB-200-2011} \cite{CUB_200_2011}, \emph{NA-Birds}~\cite{NABirds}, and \emph{Birdsnap} \cite{Birdsnap}, since this is the most challenging domain when considering current state-of-the-art results with accuracies of around \pcent{90} or less (see Table~\ref{tab:sota_results_birds}).
For other fine-grained domains like aircraft, cars, or flowers, the methods already achieve accuracies above \pcent{95}.
The three bird datasets contain between \num{200} and \num{555} different species.
CUB-200-2011 is the most popular fine-grained dataset for benchmarking because of its balanced sample distribution, but it is also the smallest one with only \num{5994} training and \num{5794} test images.
The other two datasets are more imbalanced but contain much more training images: \num{23929} and \num{40871}, respectively.
Besides class labels, bounding boxes and part annotations are available for all three datasets.

Additionally, we evaluate our method on datasets for dogs and moths species to show the applicability of our approach for other domains.
\emph{Stanford Dogs} \cite{StanfordDogs} consists of \num{120} classes and \num{20580} images with class labels and bounding box annotations.
Since the entire dataset is part of the ImageNet dataset~\cite{ImageNet}, we only use neural networks pre-trained on the iNaturalist 2017 dataset~\cite{iNaturalist} to avoid pre-training on the test images.
The \emph{EU-Moths} dataset\footnoteref{foot:link} contains \num{200} moth species common in Central Europe.
Each of the species is represented by approximately \num{11} images.
The insects are photographed manually and mainly on a relatively homogeneous background.
We manually annotated bounding boxes for each specimen and used the cropped images for training.
We trained the CNN on a random balanced split of \num{8} training and \num{3} test images per class.

\subsection{Implementation Details}
\label{sub:res:implementation}
As a primary backbone of the presented method, we take the InceptionV3 CNN architecture~\cite{Szegedy_2016_CVPR}.
We use the pre-trained weights proposed by Cui \etal~\cite{Cui_2018_CVPR_large}.
They have pre-trained the network on the iNaturalist 2017 dataset~\cite{iNaturalist} and could show that this is more beneficial for animal datasets than pre-training on ImageNet.
For some experiments~(Sect.~\ref{sub:quality_of_the_estimated_gmm_parameters} and~\ref{sub:res:ablation_study}), we also use a ResNet-50 CNN architecture~\cite{he2016deep} pre-trained on the ImageNet dataset~\cite{ILSVRC15}.

For a fair comparison, we use fixed hyperparameters for every experiment.
We train each model for 60 epochs with an AdamW optimizer~\cite{AdamW}, setting the learning rate to \num{2e-3} and $\alpha$ to \num{0.01}.
Due to limited GPU memory, we apply the gradient accumulation technique.
We use a batch size of \num{12} and accumulate the gradients over four training iterations before we perform a weight update, which results in an effective batch size of \num{48}.
Furthermore, we repeat each experiment at least \num{5} times to observe the significance and robustness of the presented approach.
The source code for our approach is publicly available on GitHub\footnote{\label{foot:link}\url{https://github.com/cvjena/fve_experiments}}.

\subsection{Fine-grained Classification}
\label{sub:res:fgvc}

%!TEX root = ../../main.tex

\begin{table}[t]
\caption{Comparison of our proposed FVE for part features with various state-of-the-art methods on three bird datasets (\textbf{bold}~=~best per dataset).}
% \todo[inline]{for EU-Moths we do not have other works (except the CS4BioDiv-WS paper, but we cannot cite it yet, right?). perhaps, we can drop these results?}
\begin{center}
	\begin{tabular}{
		 @{\hspace{0.5em}}
		l@{\hspace{2em}}
		c@{\hspace{2em}}
		c@{\hspace{2em}}
		% c@{\hspace{0.9em}}
		% c@{\hspace{0.9em}}
		c@{\hspace{0.5em}}
	}
		\toprule
			 \small{\textsc{Method}}
			 & \small{\textsc{CUB-200-2011}}
			 & \small{\textsc{NA-Birds}}
			 & \small{\textsc{Birdsnap}}
			 % & \small{\textsc{Stanford Dogs}}
			 % & \small{\textsc{EU-Moths}}
			 \\
		\midrule
			Cui \etal~\scriptsize{\cite{Cui_2018_CVPR_large}}
				& 89.3
				& 87.9
				& --
				% & 78.5
				% & --
				\\

			Stacked LSTM~\scriptsize{\cite{ge2019weakly}}
				& 90.3
				& --
				& --
				% & \emph{93.9}\scriptsize{\ding{79}}
				% & --
				\\

			FixSENet-154~\scriptsize{\cite{touvron2019fixing}}
				& 88.7
				& 89.2
				& 84.3
				% & --
				% & --
				\\

			CS-Parts~\scriptsize{\cite{Korsch19_CSPARTS}}
				& 89.5
				& 88.5
				& --
				% & --
				% & --
				\\

			MGE-CNN~\scriptsize{\cite{zhang2019learning}}
				& 89.4
				& 88.6
				& --
				% & --
				% & --
				\\

			WS-DAN~\scriptsize{\cite{hu2019see}}
				& 89.4
				& --
				& --
				% & \emph{92.2}\scriptsize{\ding{79}}
				% & --
				\\

			% DATL~\scriptsize{\cite{imran2020domain}} (\scriptsize{with~\cite{hu2019see}})
			% 	& \textbf{91.2}
			% 	& --
			% 	& --
			% 	% & 79.1
			% 	% & --
			% 	\\

			PAIRS~\scriptsize{\cite{guo2019aligned}}
				& 89.2
				& 87.9
				& --
				% & --
				% & --
				\\

			API-Net~\scriptsize{\cite{zhuang2020learning}}
				& 90.0
				& 88.1
				& --
				% & \emph{90.3}\scriptsize{\ding{79}}
				% & --
				\\

		\midrule
			% \multicolumn{2}{l}{Our classifier}
			% 	&  &  &  \\
			No Parts \scriptsize{(baseline)}
				& 89.5 \scriptsize{$\pm 0.2$}
				& 86.9 \scriptsize{$\pm 0.1$}
				& 81.9 \scriptsize{$\pm 0.5$}
				% & 77.5 \scriptsize{$\pm 0.5$}
				% & 90.5 \scriptsize{$\pm 0.5$}
				\\

			GAP \scriptsize{(parts of \cite{Korsch19_CSPARTS})}
				& 90.9 \scriptsize{$\pm 0.1$}
				& 89.9 \scriptsize{$\pm 0.1$}
				& 84.0 \scriptsize{$\pm 0.2$}
				% & 77.8 \scriptsize{$\pm 0.4$}
				% & 91.0 \scriptsize{$\pm 0.5$}
				\\

			Gradient-based FVE \scriptsize{(parts of \cite{Korsch19_CSPARTS})}
				& \textbf{91.2 \scriptsize{$\pm 0.3$}}
				& \textbf{90.4 \scriptsize{$\pm 0.1$}}
				& \textbf{85.3 \scriptsize{$\pm 0.2$}}
				% & \textbf{79.3 \scriptsize{$\pm 0.0$}*}
				% & \textbf{93.0 \scriptsize{$\pm 1.2$}}
				\\

			EM-based FVE \scriptsize{(parts of \cite{Korsch19_CSPARTS})}
				& \textbf{91.1 \scriptsize{$\pm 0.2$}}
				& \textbf{90.3 \scriptsize{$\pm 0.1$}}
				& \textbf{84.9 \scriptsize{$\pm 0.2$}}
				% & \textbf{79.2 \scriptsize{$\pm 0.1$}}
				% & \textbf{92.0 \scriptsize{$\pm 0.9$}}
				\\
		\bottomrule
	\end{tabular}
	\label{tab:sota_results_birds}
% 	\vspace{-0.5cm}
\end{center}
\end{table}

\begin{table}[t]
	\caption{Results on the Stanford Dogs and EU-Moths datasets. For Stanford Dogs, we only compare to methods that do not use ImageNet pre-training. Similar to our work, they utilize a pre-training on the iNaturalist 2017 dataset. This kind of pre-training results in a more fair comparison, since the training set of ImageNet contains the test set of Stanford Dogs.}

	\begin{center}

	\begin{tabular}{
		 @{\hspace{0.5em}}
		l@{\hspace{2em}}
		c@{\hspace{2em}}
		c@{\hspace{0.5em}}
	}

	\toprule
		 \small{\textsc{Method}}
		 & \small{\textsc{Stanford Dogs}}
		 & \small{\textsc{EU-Moths}}
		 \\

	\midrule
		Cui \etal~\scriptsize{\cite{Cui_2018_CVPR_large}}
			& 78.5
			& --
			\\

		DATL~\scriptsize{\cite{imran2020domain}} (\scriptsize{with~\cite{hu2019see}})
			& 79.1
			& --
			\\
	\midrule
		No Parts \scriptsize{(baseline)}
			& 77.5 \scriptsize{$\pm 0.5$}
			& 90.5 \scriptsize{$\pm 0.5$}

			\\

		GAP \scriptsize{(parts of \cite{Korsch19_CSPARTS})}
			& 77.8 \scriptsize{$\pm 0.4$}
			& 91.0 \scriptsize{$\pm 0.5$}
			\\

		Gradient-based FVE \scriptsize{(parts of \cite{Korsch19_CSPARTS})}
			& \textbf{79.1 \scriptsize{$\pm 0.3$}}
			& \textbf{93.0 \scriptsize{$\pm 1.2$}}
			\\

		EM-based FVE \scriptsize{(parts of \cite{Korsch19_CSPARTS})}
			& \textbf{79.2 \scriptsize{$\pm 0.1$}}
			& \textbf{92.0 \scriptsize{$\pm 0.9$}}
			\\
	\bottomrule
	\end{tabular}
	\end{center}
	\label{tab:sota_results_dogs_moths}
% 	\vspace{-0.4cm}
\end{table}

In our first experiment, we test our proposed FVE-Layer together with an unsupervised part detector that provides classification-specific parts (CS-Parts)~\cite{Korsch19_CSPARTS}.
However, in contrast to Korsch \etal~\cite{Korsch19_CSPARTS}, we use a separate CNN for calculating part features.
This part-CNN is fine-tuned on the detected parts, and the extracted features are adapted to the FVE.
Besides the part-CNN, we also extract features from the global image with another CNN.
For the part-CNN, the prediction is performed based on the FVE of the part features, whereas the prediction on the global image is made based on standard CNN features.
Both predictions are then weighted equally and summed up to the final prediction.
The GMM parameters are either estimated via gradient descent or by our proposed online EM algorithm.
After investigating the effect of both number and dimension of the mixture components (see supplementary material), we use one component with a dimension of \num{2048}.
We also compare this approach with results based on concatenated part features and GAP, and with a baseline using only the prediction obtained from the global image (no parts).

For bird species classification, Table~\ref{tab:sota_results_birds} contains our results as well as accuracies reported in previous work.
Our proposed FVE for part features performs best on all three datasets.
The results for NA-Birds stand out, since our approach is the only one reaching an accuracy greater than \pcent{90} on this challenging dataset
Accuracies for dogs and moth species are shown in Table~\ref{tab:sota_results_dogs_moths}.
Again, our FVE approach performs best, showing its suitability beyond the bird species domain.

\subsection{Quality of the Estimated GMM Parameters} % (fold)
\label{sub:quality_of_the_estimated_gmm_parameters}
We now investigate the quality of the estimated GMM parameters with respect to the two proposed approaches (gradient-based vs. EM-based).
For this purpose, we first compare them on a generated dataset consisting of feature vectors for five classes sampled from different normal distributions.
The corresponding mean vectors (class centroids) are arranged on the unit sphere, and we set the variance of the class distributions to $\frac{1}{5}$ such that the features do not overlap.
We trained a simple neural network consisting of the FVE-Layer and a linear layer as a classifier.
The data dimension has been varied, using powers of two in the range $2$ to $256$, which also defines the dimensions of the mixture components.
Each setup was repeated five times.
As a baseline, we also estimated the GMM parameters with a standard EM algorithm independent of neural network training. %in an offline manner.

%!TEX root = ../../main.tex

\begin{figure}[t]
	\begin{center}
	\begin{subfigure}[t]{0.49\textwidth}
		\includegraphics[width=\linewidth]{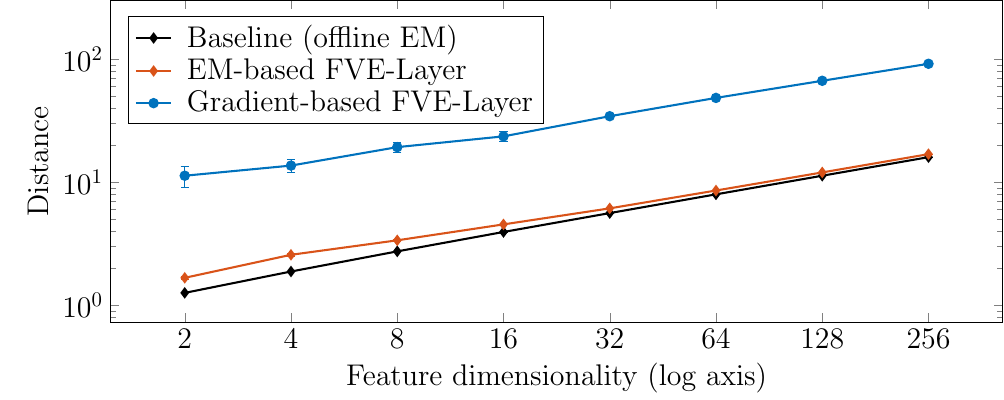}
		\caption{One component}
		\label{fig:distances:1_comp}
	\end{subfigure}
	\hfill
	\begin{subfigure}[t]{0.49\textwidth}
		\includegraphics[width=\linewidth]{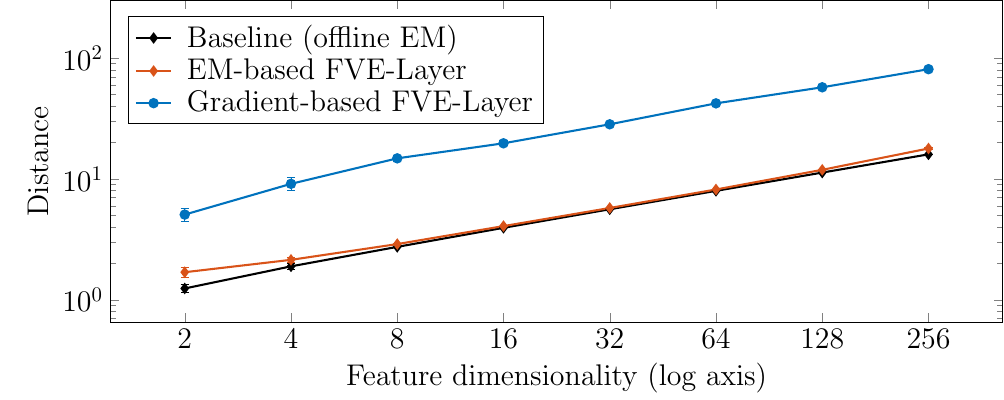}
		\caption{Five components}
		\label{fig:distances:5_comp}
	\end{subfigure}
	\caption{Comparison of the weighted normalized Euclidean distance (see Eq.~\eqref{eq:euc_distance}) on a generated dataset. The dataset consist of five classes and the experiment was performed with one and five GMM components.}
	\label{fig:distances}
	\vspace{-0.5cm}
	\end{center}
\end{figure}

In Figure~\ref{fig:distances}, we visualize the normalized Euclidean distance of the estimated parameters averaged over the entire dataset.
The normalized Euclidean distance for a single feature vector can be computed via:
\begin{equation}
D(\vec{x}_{n,t}|\theta) = \sum_k w_{n,t,k} \cdot \sqrt{\sum_d\frac{\left(x_{n,t,d} - \mu_{k,d}\right)^2}{\left(\sigma_{k,d}\right)^2}} \quad . \label{eq:euc_distance}
\end{equation}
It is the sum of distances to the mean vectors of the mixture components, normalized by the corresponding variances and weighted by the soft-assignments $w_{n,t,k}$.
Furthermore, we estimate the same distances for the CUB-200-2011 dataset, shown in Table~\ref{tab:2D_distance}.
Both evaluations on synthetic and real data show that GMMs estimated by our proposed online EM algorithm fit the data much better, resulting in more precise clusters due to lower normalized Euclidean distances.
Moreover, we show in the supplementary material that the gradient-based method changes the GMM parameters only slightly for two-dimensional data, whereas the online EM algorithm estimates parameters as good as the offline EM algorithm such that the mixtures match the data distributions well.

%!TEX root = ../../main.tex

\begin{table}[t]
\caption{Comparison of weighted normalized Euclidean distances (see Eq.~\eqref{eq:euc_distance}) for the three birds datasets, evaluated for two CNN architectures.}
\begin{center}
	\begin{tabular}{
		@{\hspace{0.5em}}l@{\hspace{0.5em}}
		@{\hspace{0.7em}}cc@{\hspace{0.7em}}
		@{\hspace{0.7em}}cc@{\hspace{0.7em}}
	}

		\toprule

			& \multicolumn{2}{@{\hspace{0.7em}}c@{\hspace{0.7em}}}{\scriptsize{\textsc{GMM Estimation for ResNet50}}}
			& \multicolumn{2}{@{\hspace{0.7em}}c@{\hspace{0.7em}}}{\scriptsize{\textsc{GMM Estimation for InceptionV3}}}
			\\

			& \scriptsize{\textsc{EM-based}}
			& \scriptsize{\textsc{Gradient-based}}
			& \scriptsize{\textsc{EM-based}}
			& \scriptsize{\textsc{Gradient-based}}
			\\

		\midrule

		\small{\textsc{CUB-200-2011}}
			& 21.49 \scriptsize{($\pm 0.68$)}
			& 46.85 \scriptsize{($\pm 0.18$)}
			& 33.21 \scriptsize{($\pm 0.13$)}
			& 37.15 \scriptsize{($\pm 0.18$)}
			\\

		\small{\textsc{NA-Birds}}
			& 18.36 \scriptsize{($\pm 1.00$)}
			& 45.35 \scriptsize{($\pm 0.18$)}
			& 36.62 \scriptsize{($\pm 0.38$)}
			& 35.50 \scriptsize{($\pm 0.15$)}
			\\

		\small{\textsc{Birdsnap}}
			& 11.08 \scriptsize{($\pm 2.01$)}
			& 46.75 \scriptsize{($\pm 0.23$)}
			& 32.75 \scriptsize{($\pm 0.31$)}
			& 35.62 \scriptsize{($\pm 0.30$)}
			\\

		\bottomrule
		\end{tabular}
		\label{tab:2D_distance}
\end{center}
\vspace{-8pt}
\end{table}

\subsection{Ablation Study} % (fold)
\label{sub:res:ablation_study}
% In this experiment, we investigate the impact of the different stages of the proposed method.
In an ablation study, we investigate the impact of the FVE in the proposed approach.
As seen in Figure~\ref{fig:method}, our method consists of two branches: classification of the global image and classification based on estimated parts.
In Table~\ref{tab:ablation_study}, we show accuracies achieved by the individual branches as well as by the combined classification for two CNN architectures.
We see that part features with FVE result in better classification accuracies compared to GAP.
This effect propagates to the final accuracy, resulting in improved classification performance.

%!TEX root = ../../main.tex

\begin{table}[t]
\caption{Ablation study on the CUB-200-2011 dataset with two different CNNs.}
\begin{center}
	\begin{tabular}{@{\hspace{0.5em}}l@{\hspace{2.5em}}c@{\hspace{2.5em}}c@{\hspace{0.5em}}}
		\toprule
			\small{\textsc{Method}} &
			\small{\textsc{ResNet50}}  &
			\small{\textsc{InceptionV3}}
			\\
		\midrule
			\small{\textsc{Baseline CNN}}
			& 84.4 \scriptsize{$\pm 0.3$}
			& 89.5 \scriptsize{$\pm 0.2$}
			\\
		\midrule
			\small{\textsc{Parts \scriptsize{\cite{Korsch19_CSPARTS}} + GAP}}
			& 79.7 \scriptsize{$\pm 0.3$}
			& 88.8 \scriptsize{$\pm 0.2$}
			\\
			\small{\textsc{Parts \scriptsize{\cite{Korsch19_CSPARTS}} + FVE}}
			& 82.6 \scriptsize{$\pm 0.4$}
			& 89.1 \scriptsize{$\pm 0.2$}
			\\
		\midrule
			\small{\textsc{Baseline CNN + Parts \scriptsize{\cite{Korsch19_CSPARTS}} + GAP}}
			& 85.9 \scriptsize{$\pm 0.2$}
			& 90.9 \scriptsize{$\pm 0.1$}
			\\
			\small{\textsc{Baseline CNN + Parts \scriptsize{\cite{Korsch19_CSPARTS}} + FVE}}
			& 86.4 \scriptsize{$\pm 0.2$}
			& 91.1 \scriptsize{$\pm 0.2$}
			\\
		\bottomrule
		\end{tabular}
		\label{tab:ablation_study}
\end{center}
\vspace{-8pt}
\end{table}

   %!TEX root = ../main.tex

\section{Conclusions}

In this paper, we have proposed a new FVE-Layer for aggregating part features of a CNN in the context of fine-grained categorization, which uses an online EM algorithm for estimating the underlying GMM jointly with other network parameters.
With this layer, we are able to compute a unified fixed-length representation for a varying number of local part features, which allows a deep neural network to cope with missing parts as well as with an arbitrary order of part features, e.g., given by an unsupervised part detector.
% Furthermore, our proposed layer can be trained jointly with other layers requiring only minimal computational overhead.
In our experiments, we have achieved state-of-the-art recognition accuracies on three fine-grained datasets for bird species classification: CUB-200-2011 (\pcent{91.1}), NA-Birds (\pcent{90.3}), and Birdsnap (\pcent{84.9}).
Furthermore, we have shown that compared to existing deep FVE implementations, our online EM-based approach results in more accurate estimates of the mixture model.

% In our experiments, we have shown that the proposed method outperforms the conventional pipelines due to the joint learning of both part features and their encoding.
% Furthermore, the FVE-Layer is more robust compared to typical CNN architectures if the correct semantic meaning of detected parts is not present, as it is the case for unsupervised part detection.
% Finally, .

   % \section*{Acknowledgments}
   % \todo[inline]{Do we need this one here?}
   % This work has been funded by the German Federal Ministry of Education and Research (Bundesministerium für Bildung und Forschung, BMBF, Deutschland) via the project ``\mbox{Development} of an \mbox{Automated} \mbox{Multisensor} \mbox{Station} for \mbox{Monitoring} of \mbox{Biodiversity} (\mbox{AMMOD}) - Subproject 5: Automated Visual Monitoring and Analysis'' (FKZ: 01LC1903E).

   \clearpage
   \bibliographystyle{splncs04}
   \bibliography{main}

\begin{thebibliography}{10}
\providecommand{\url}[1]{\texttt{#1}}
\providecommand{\urlprefix}{URL }
\providecommand{\doi}[1]{https://doi.org/#1}

\bibitem{arandjelovic2016netvlad}
Arandjelovic, R., Gronat, P., Torii, A., Pajdla, T., Sivic, J.: Netvlad: Cnn
  architecture for weakly supervised place recognition. In: Proceedings of the
  IEEE Conference on Computer Vision and Pattern Recognition. pp. 5297--5307
  (2016)

\bibitem{SequentialEM:2009}
Awwad Shiekh~Hasan, B., Gan, J.Q.: Sequential em for unsupervised adaptive
  gaussian mixture model based classifier. In: Perner, P. (ed.) Machine
  Learning and Data Mining in Pattern Recognition. pp. 96--106. Springer Berlin
  Heidelberg, Berlin, Heidelberg (2009)

\bibitem{Birdsnap}
Berg, T., Liu, J., Woo~Lee, S., Alexander, M.L., Jacobs, D.W., Belhumeur, P.N.:
  Birdsnap: Large-scale fine-grained visual categorization of birds. In:
  Proceedings of the IEEE Conference on Computer Vision and Pattern
  Recognition. pp. 2011--2018 (2014)

\bibitem{OnlineEM:2009}
Cappe, O., Moulines, E.: On-line expectation-maximization algorithm for latent
  data models. Journal of the Royal Statistical Society. Series B (Statistical
  Methodology)  \textbf{71}(3),  593--613 (2009)

\bibitem{chen2018stochastic}
Chen, J., Zhu, J., Teh, Y.W., Zhang, T.: Stochastic expectation maximization
  with variance reduction. In: Advances in Neural Information Processing
  Systems. pp. 7967--7977 (2018)

\bibitem{cimpoi2015deep}
Cimpoi, M., Maji, S., Vedaldi, A.: Deep filter banks for texture recognition
  and segmentation. In: Proceedings of the IEEE Conference on Computer Vision
  and Pattern Recognition. pp. 3828--3836 (2015)

\bibitem{Cui_2018_CVPR_large}
Cui, Y., Song, Y., Sun, C., Howard, A., Belongie, S.: Large scale fine-grained
  categorization and domain-specific transfer learning. In: Proceedings of the
  IEEE Conference on Computer Vision and Pattern Recognition (6 2018)

\bibitem{ImageNet}
Deng, J., Dong, W., Socher, R., Li, L.J., Li, K., Fei-Fei, L.: Imagenet: A
  large-scale hierarchical image database. In: 2009 IEEE conference on computer
  vision and pattern recognition. pp. 248--255. Ieee (2009)

\bibitem{Fu_2017_CVPR}
Fu, J., Zheng, H., Mei, T.: Look closer to see better: Recurrent attention
  convolutional neural network for fine-grained image recognition. In:
  Proceedings of the IEEE Conference on Computer Vision and Pattern Recognition
  (7 2017)

\bibitem{ge2019weakly}
Ge, W., Lin, X., Yu, Y.: Weakly supervised complementary parts models for
  fine-grained image classification from the bottom up. In: Proceedings of the
  IEEE Conference on Computer Vision and Pattern Recognition. pp. 3034--3043
  (2019)

\bibitem{gong2014vlad_pooling}
Gong, Y., Wang, L., Guo, R., Lazebnik, S.: Multi-scale orderless pooling of
  deep convolutional activation features. In: Proceedings of the European
  Conference on Computer Vision. pp. 392--407. Springer (2014)

\bibitem{guo2019aligned}
Guo, P., Farrell, R.: Aligned to the object, not to the image: A unified
  pose-aligned representation for fine-grained recognition. In: IEEE Winter
  Conference on Applications of Computer Vision. pp. 1876--1885. IEEE (2019)

\bibitem{adaptiveGMM:2009}
{Hasan}, B.A.S., {Gan}, J.Q.: Unsupervised adaptive gmm for bci. In: 2009 4th
  International IEEE/EMBS Conference on Neural Engineering. pp. 295--298 (April
  2009)

\bibitem{he2016deep}
He, K., Zhang, X., Ren, S., Sun, J.: Deep residual learning for image
  recognition. In: IEEE Conference on Computer Vision and Pattern Recognition.
  pp. 770--778 (2016)

\bibitem{he2019and}
He, X., Peng, Y., Zhao, J.: Which and how many regions to gaze: Focus
  discriminative regions for fine-grained visual categorization. International
  Journal of Computer Vision pp. 1--21 (2019)

\bibitem{hu2019see}
Hu, T., Qi, H., Huang, Q., Lu, Y.: See better before looking closer: Weakly
  supervised data augmentation network for fine-grained visual classification.
  arXiv preprint arXiv:1901.09891  (2019)

\bibitem{imran2020domain}
Imran, A., Athitsos, V.: Domain adaptive transfer learning on visual attention
  aware data augmentation for fine-grained visual categorization. In:
  International Symposium on Visual Computing. pp. 53--65. Springer (2020)

\bibitem{ioffe2015batch}
Ioffe, S., Szegedy, C.: Batch normalization: Accelerating deep network training
  by reducing internal covariate shift. arXiv preprint arXiv:1502.03167  (2015)

\bibitem{jaakkola1999exploiting}
Jaakkola, T., Haussler, D.: Exploiting generative models in discriminative
  classifiers. In: Advances in Neural Information Processing Systems. pp.
  487--493 (1999)

\bibitem{jegou2010aggregating}
J{\'e}gou, H., Douze, M., Schmid, C., P{\'e}rez, P.: Aggregating local
  descriptors into a compact image representation. In: 2010 IEEE computer
  society conference on computer vision and pattern recognition. pp.
  3304--3311. IEEE (2010)

\bibitem{karimi2019global}
Karimi, B., Wai, H.T., Moulines, {\'E}., Lavielle, M.: On the global
  convergence of (fast) incremental expectation maximization methods. In:
  Advances in Neural Information Processing Systems. pp. 2833--2843 (2019)

\bibitem{StanfordDogs}
Khosla, A., Jayadevaprakash, N., Yao, B., Li, F.F.: Novel dataset for
  fine-grained image categorization: Stanford dogs. In: Proc. CVPR Workshop on
  Fine-Grained Visual Categorization (FGVC). vol.~2. Citeseer (2011)

\bibitem{kingma2014adam}
Kingma, D.P., Ba, J.: Adam: A method for stochastic optimization. arXiv
  preprint arXiv:1412.6980  (2014)

\bibitem{Korsch19_CSPARTS}
Korsch, D., Bodesheim, P., Denzler, J.: Classification-specific parts for
  improving fine-grained visual categorization. In: Proceedings of the German
  Conference on Pattern Recognition. pp. 62--75 (2019)

\bibitem{krause2016unreasonable}
Krause, J., Sapp, B., Howard, A., Zhou, H., Toshev, A., Duerig, T., Philbin,
  J., Fei-Fei, L.: The unreasonable effectiveness of noisy data for
  fine-grained recognition. In: Proceedings of the European Conference on
  Computer Vision. pp. 301--320. Springer (2016)

\bibitem{liang2009onlineEM}
Liang, P., Klein, D.: Online em for unsupervised models. In: Proceedings of
  human language technologies: The 2009 annual conference of the North American
  chapter of the association for computational linguistics. pp. 611--619 (2009)

\bibitem{lin2015bilinear}
Lin, T.Y., RoyChowdhury, A., Maji, S.: Bilinear cnn models for fine-grained
  visual recognition. In: Proceedings of the IEEE International Conference on
  Computer Vision. pp. 1449--1457 (2015)

\bibitem{AdamW}
Loshchilov, I., Hutter, F.: Decoupled weight decay regularization. arXiv
  preprint arXiv:1711.05101  (2017)

\bibitem{neal1998view}
Neal, R.M., Hinton, G.E.: A view of the em algorithm that justifies
  incremental, sparse, and other variants. In: Learning in Graphical Models,
  pp. 355--368. Springer (1998)

\bibitem{perronnin2007fisher}
Perronnin, F., Dance, C.: Fisher kernels on visual vocabularies for image
  categorization. In: Proceedings of the IEEE Conference on Computer Vision and
  Pattern Recognition. pp.~1--8. IEEE (2007)

\bibitem{Perronnin10:ITF}
{Perronnin}, F., {Sanchez}, J., {Mensink}, T.: Improving the fisher kernel for
  large-scale image classification. In: Proceedings of the European Conference
  on Computer Vision. Lecture Notes in Computer Science, vol.~6314, pp.
  143--156. Springer (2010)

\bibitem{ILSVRC15}
Russakovsky, O., Deng, J., Su, H., Krause, J., Satheesh, S., Ma, S., Huang, Z.,
  Karpathy, A., Khosla, A., Bernstein, M., Berg, A.C., Fei-Fei, L.: {ImageNet
  Large Scale Visual Recognition Challenge}. International Journal of Computer
  Vision  \textbf{115}(3),  211--252 (2015). \doi{10.1007/s11263-015-0816-y}

\bibitem{Simon17_GOP}
Simon, M., Gao, Y., Darrell, T., Denzler, J., Rodner, E.: Generalized orderless
  pooling performs implicit salient matching. In: Proceedings of the IEEE
  International Conference on Computer Vision. pp. 4970--4979 (2017)

\bibitem{Simon19:Implicit}
Simon, M., Rodner, E., Darell, T., Denzler, J.: The whole is more than its
  parts? from explicit to implicit pose normalization. IEEE Transactions on
  Pattern Analysis and Machine Intelligence pp. 1--13 (2018)

\bibitem{simonyan2013fve_layer}
Simonyan, K., Vedaldi, A., Zisserman, A.: Deep fisher networks for large-scale
  image classification. In: Advances in Neural Information Processing Systems.
  pp. 163--171 (2013)

\bibitem{song2017locally}
Song, Y., Zhang, F., Li, Q., Huang, H., O'Donnell, L.J., Cai, W.:
  Locally-transferred fisher vectors for texture classification. In:
  Proceedings of the IEEE International Conference on Computer Vision. pp.
  4912--4920 (2017)

\bibitem{sydorov2014deep}
Sydorov, V., Sakurada, M., Lampert, C.H.: Deep fisher kernels-end to end
  learning of the fisher kernel gmm parameters. In: Proceedings of the IEEE
  Conference on Computer Vision and Pattern Recognition. pp. 1402--1409 (2014)

\bibitem{szegedy2015going}
Szegedy, C., Liu, W., Jia, Y., Sermanet, P., Reed, S., Anguelov, D., Erhan, D.,
  Vanhoucke, V., Rabinovich, A.: Going deeper with convolutions. In:
  Proceedings of the IEEE conference on computer vision and pattern
  recognition. pp.~1--9 (2015)

\bibitem{Szegedy_2016_CVPR}
Szegedy, C., Vanhoucke, V., Ioffe, S., Shlens, J., Wojna, Z.: Rethinking the
  inception architecture for computer vision. In: Proceedings of the IEEE
  Conference on Computer Vision and Pattern Recognition (June 2016)

\bibitem{tang2018deep}
Tang, P., Wang, X., Shi, B., Bai, X., Liu, W., Tu, Z.: Deep fishernet for image
  classification. IEEE Transactions on Neural Networks and Learning Systems
  \textbf{30}(7),  2244--2250 (2018)

\bibitem{touvron2019fixing}
Touvron, H., Vedaldi, A., Douze, M., J{\'e}gou, H.: Fixing the train-test
  resolution discrepancy. In: Advances in Neural Information Processing
  Systems. pp. 8250--8260 (2019)

\bibitem{NABirds}
{Van Horn}, G., {Branson}, S., {Farrell}, R., {Haber}, S., {Barry}, J.,
  {Ipeirotis}, P., {Perona}, P., {Belongie}, S.: Building a bird recognition
  app and large scale dataset with citizen scientists: The fine print in
  fine-grained dataset collection. In: Proceedings of the IEEE Conference on
  Computer Vision and Pattern Recognition. pp. 595--604 (June 2015)

\bibitem{iNaturalist}
Van~Horn, G., Mac~Aodha, O., Song, Y., Cui, Y., Sun, C., Shepard, A., Adam, H.,
  Perona, P., Belongie, S.: The inaturalist species classification and
  detection dataset. In: Proceedings of the IEEE Conference on Computer Vision
  and Pattern Recognition. pp. 8769--8778 (2018)

\bibitem{CUB_200_2011}
Wah, C., Branson, S., Welinder, P., Perona, P., Belongie, S.: The caltech-ucsd
  birds-200-2011 dataset. Tech. Rep. CNS-TR-2011-001, California Institute of
  Technology (2011)

\bibitem{wieschollek2017backpropagation}
Wieschollek, P., Groh, F., Lensch, H.: Backpropagation training for fisher
  vectors within neural networks. arXiv preprint arXiv:1702.02549  (2017)

\bibitem{yang2021rerank}
Yang, S., Liu, S., Yang, C., Wang, C.: Re-rank coarse classification with local
  region enhanced features for fine-grained image recognition. arXiv preprint
  arXiv:2102.09875  (2021)

\bibitem{yang2018_NTSNet}
Yang, Z., Luo, T., Wang, D., Hu, Z., Gao, J., Wang, L.: Learning to navigate
  for fine-grained classification. In: Proceedings of the European Conference
  on Computer Vision (ECCV). pp. 420--435 (2018)

\bibitem{zhang2019unsupervised}
Zhang, J., Zhang, R., Huang, Y., Zou, Q.: Unsupervised part mining for
  fine-grained image classification. arXiv preprint arXiv:1902.09941  (2019)

\bibitem{zhang2019learning}
Zhang, L., Huang, S., Liu, W., Tao, D.: Learning a mixture of
  granularity-specific experts for fine-grained categorization. In: Proceedings
  of the IEEE International Conference on Computer Vision. pp. 8331--8340
  (2019)

\bibitem{zhang2016picking}
Zhang, X., Xiong, H., Zhou, W., Lin, W., Tian, Q.: Picking deep filter
  responses for fine-grained image recognition. In: Proceedings of the IEEE
  Conference on Computer Vision and Pattern Recognition. pp. 1134--1142 (2016)

\bibitem{zheng2017learning}
Zheng, H., Fu, J., Mei, T., Luo, J.: Learning multi-attention convolutional
  neural network for fine-grained image recognition. In: Proceedings of the
  IEEE International Conference on Computer Vision (2017)

\bibitem{zheng2019learning}
Zheng, H., Fu, J., Zha, Z.J., Luo, J.: Learning deep bilinear transformation
  for fine-grained image representation. In: Advances in Neural Information
  Processing Systems. pp. 4279--4288 (2019)

\bibitem{zhuang2020learning}
Zhuang, P., Wang, Y., Qiao, Y.: Learning attentive pairwise interaction for
  fine-grained classification. In: Proceedings of the AAAI Conference on
  Artificial Intelligence. vol.~34, pp. 13130--13137 (2020)

\end{thebibliography}
\end{document}